\documentclass[10pt,twocolumn,letterpaper]{article}

\usepackage{cvpr}              

\usepackage{microtype}

\renewcommand{\paragraph}[1]{\vspace{.5em}\noindent\textbf{#1.}}

\usepackage{algorithm}
\usepackage{algorithmic}
\definecolor{commentcolor}{RGB}{69, 6, 147}
\definecolor{rowcolor}{RGB}{229, 217, 242}
\usepackage{caption}
\usepackage{booktabs}
\usepackage{multirow}
\usepackage{color, colortbl}
\usepackage{subcaption}
\usepackage{varwidth}
\usepackage[most]{tcolorbox}
\newcommand\fire{\raisebox{-0.1em}{\includegraphics[width=1em]{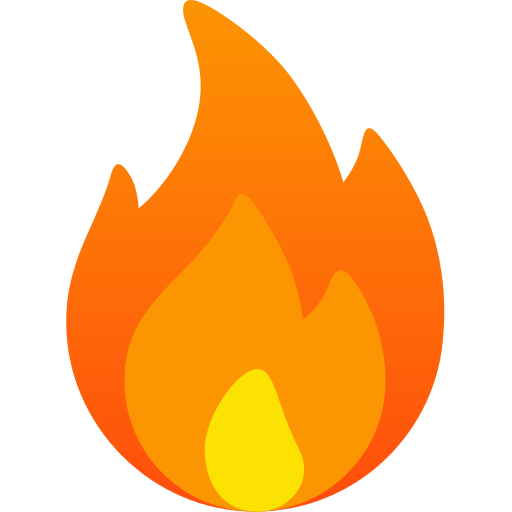}}}
\newcommand\snow{\raisebox{-0.2em}{\includegraphics[width=1em]{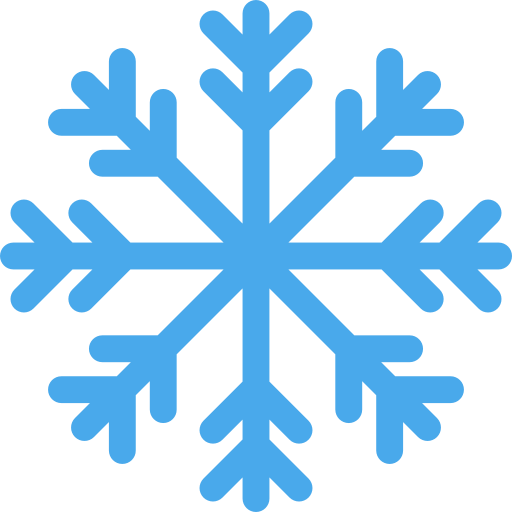}}}
\def\thickhline{\noalign{\hrule height2.0pt}}

\definecolor{cvprblue}{rgb}{0.21,0.49,0.74}
\definecolor{correctgreen}{RGB}{65, 166, 126}
\definecolor{citecolor2}{RGB}{186, 72, 127}
\usepackage[pagebackref,breaklinks,colorlinks,allcolors=cvprblue]{hyperref}

\def\paperID{***} 
\def\confName{CVPR}
\def\confYear{2026}

\title{Don’t Show Pixels, Show Cues: \\ Unlocking Visual Tool Reasoning in Language Models via Perception Programs}

\author{Muhammad Kamran Janjua\(^{1*}\)~~Hugo Silva\(^{1*}\)~~Di Niu\(^2\)~~Bahador Rashidi\(^1\)\\
\(^{1}\)Huawei Technologies, Canada~~\(^2\)University of Alberta, Canada\\
\href{https://github.com/AISmartPerception/perception-programs}{\texttt{\textcolor{citecolor2}{github.com/AISmartPerception/perception-programs}}}
}

\begin{document}
\twocolumn[{%
\renewcommand\twocolumn[1][]{#1}%
\maketitle
\begin{center}
    \centering
    \captionsetup{type=figure}
    \includegraphics[width=0.9\textwidth]{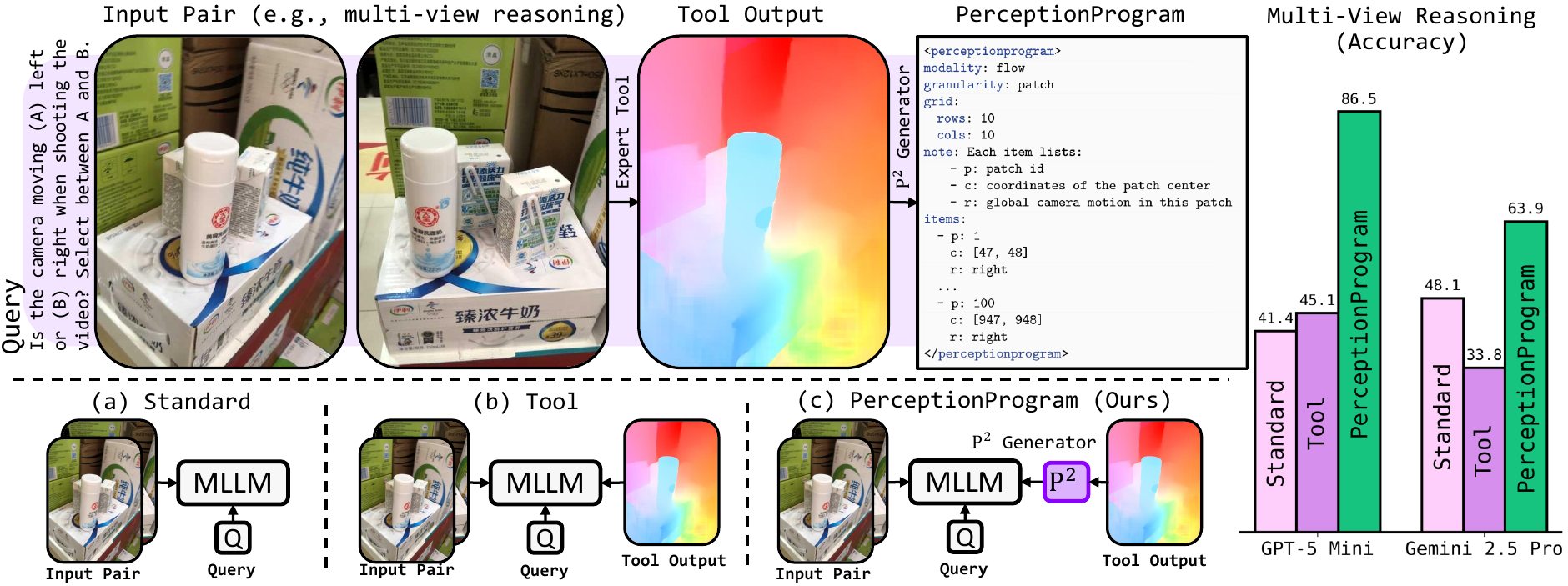}
    \captionof{figure}{\textbf{Teaser.} Turning dense tool outputs into a Perception Program makes a general MLLM behave as if it can \textit{read} the modality. Given same query and input pair, (a) standard MLLM underuses the visual signal~\citep{fu2025hiddenplainsightvlms}, (b) a tool-only route exposes the modality but stays pixel-level, while (c) our \textsc{P}\(^2\) summarizes it into a language-native structure that MLLM can reliably reason over, yielding large gains.}
    \label{fig:teaser}
\end{center}
}]
\def\thefootnote{*}\footnotetext{Equal Contribution \\ correspondence: \tt\small muhammad.kamran.janjua@huawei.com}

\begin{abstract}
Multimodal language models (MLLMs) are increasingly paired with vision tools (e.g., depth, flow, correspondence) to enhance visual reasoning. However, despite access to these tool-generated visual cues, MLLMs often fail to benefit fully from them. Existing approaches typically feed raw tool outputs into the model, but these dense, pixel-level representations are misaligned with the language-native reasoning strengths of LLMs, leading to weak perception due to reliance on language priors. We argue that, in problems where vision tools can provide the necessary visual cues, the bottleneck is not more tool calls or larger MLLMs, it is how tool outputs are represented. We introduce Perception Programs (\textsc{P}\(^2\)), a training-free, model-agnostic method that rewrites tool outputs into compact, structured, language-native summaries that MLLMs can directly parse and reason over.
Across six perception-centric tasks in BLINK, \textsc{P}\(^2\) consistently yields large improvements over base models and raw tool-augmented baselines. With GPT-5 Mini as the base model, \textsc{P}\(^2\) raises its accuracy from \(41.35\%\) to \(86.47\%\) on multi-view reasoning, from \(52.42\%\) to \(81.45\%\) on relative depth, and achieves a \(19.66\%\) overall average gain across tasks, setting new state-of-the-art results. Even on smaller MLLMs, e.g., InternVL3.5-\(4\)B and Qwen3VL-\(4\)B, we observe \(21\)–\(25\%\) absolute gains from \textsc{P}\(^2\) across tasks when comparing to image-based tool variants, surpassing prior agentic, supervised, and RL-based tool-use methods, without any training or model modifications.
\end{abstract}    
\section{Introduction}
\label{sec:intro}

\begin{figure}[!t]
    \centering
    \includegraphics[width=\linewidth]{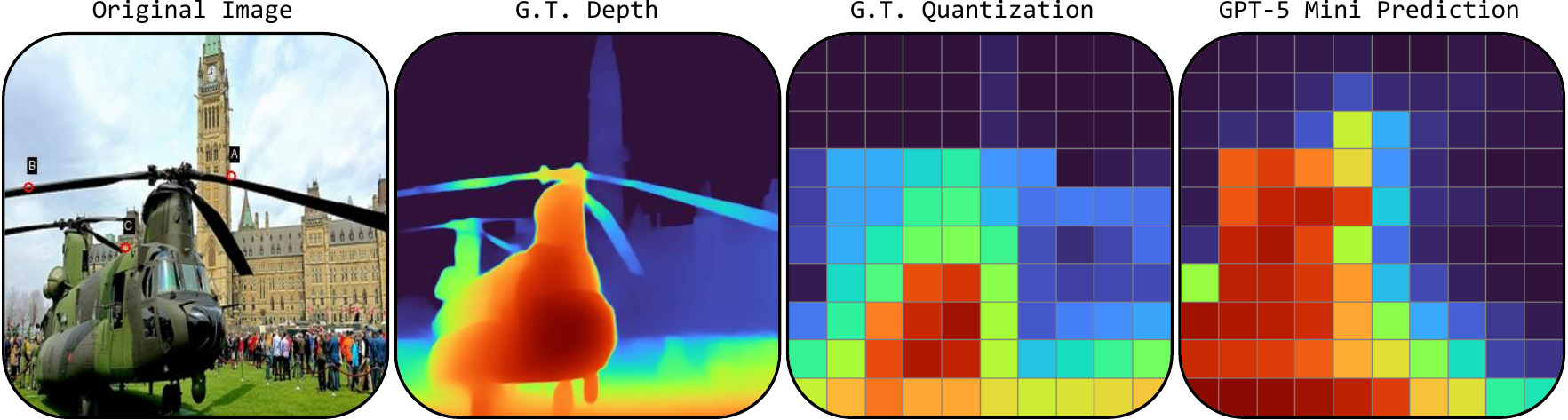}
    \caption{\textbf{Under-utilization of Visual Information.} Given several ICL examples along with depth map, GPT-5 Mini fails to recover near-to-far ordering from it (see \cref{subsec:analysis}), indicating limited utilization of the modality.}
    \label{fig:gpt5illustration}
\end{figure}

Modern multimodal language models (MLLMs) are increasingly expected to perform perception-driven reasoning over visual inputs such as images and video~\citep{fu2024blink,patraucean2023perception}. Recent advances have made it feasible to pair MLLMs with vision tools, e.g., monocular depth estimation, optical flow, visual correspondence, and object detectors, to surface perceptual signals that are not directly accessible from pixel inputs alone~\citep{zeng2022socratic,bigverdi2025perception,yang2025machine}. Despite this, MLLMs often under-utilize the rich information provided by these tools. When raw tool outputs are serialized and supplied to the model, they appear as dense, low-level visual tokens that misalign with the language-native reasoning substrate of LLMs~\citep{fu2025hiddenplainsightvlms}. As a result, MLLMs exhibit limited perceptual grounding, inherit language priors from the base LLM, and frequently fail on tasks that require interpreting visual modalities, even when provided with accurate tool surrogates for those modalities, see~\cref{fig:gpt5illustration} for an illustration.

Prior work has attempted to mitigate this limitation through (i) program-synthesis pipelines that generate executable code to invoke tools~\citep{suris2023vipergpt,gupta2023visual}, (ii) chain-of-thought-based tool-use methods that interleave tool calls with reasoning~\citep{bigverdi2025perception,yang2025machine}, (iii) fine-tuning (SFT or reinforcement-learning) to encourage tool-augmented reasoning~\citep{zhou2025reinforced,sarch2025grounded,ma2025latte,zhang2025thyme,yu2025introducing}, or (iv) architectures incorporating perception-oriented modules~\citep{tang2025tulip,cho2025perceptionlm}. These approaches, however, increase computational cost, require specialized training, or continue to operate at the same pixel-level granularity as the raw tool outputs, thus inheriting the fundamental representation mismatch between visual tool outputs and the linguistic reasoning capabilities of MLLMs. Collectively, these trends raise a central question: 

\begin{tcolorbox}[colback=gray!5!white,colframe=black!75!black]
  \textit{Is the true bottleneck in tool-augmented MLLMs the number of tool calls or model size—or the representation of the visual information itself?}
\end{tcolorbox}


The main motivation of this work stems from contrasting how visual information is fundamentally captured by humans with the pixel-level processing often imposed on MLLMs. Human cue-taking varies widely depending on the type of data: surface proximity for depth-related problems, left–right movement for spatial orientation, or local similarity when looking for correspondences between images. Converting the key information in each of these problems into text reduces the models’ burden of processing pixel-level details and better equips them to interpret and reason about such information, since text is the representation that best aligns with their native reasoning capabilities. 

With that in mind, we introduce Perception Programs (\textsc{P}\(^2\)), a training-free, model-agnostic representation that rewrites raw tool outputs into compact, structured, language-native summaries of visual modalities. \textsc{P}\(^2\) standardizes what is conveyed from a tool (the perceptual signal), where it is grounded spatially in the image, and (optionally) how local regions relate. This reformulation enables MLLMs to read the visual modality rather than infer it from dense numeric tokens. \textsc{P}\(^2\) requires no parameter updates to the MLLM, no architectural modifications, and no additional tool calls during inference. It serves as a plug-and-play module: the same tool output provided to a standard tool-use pipeline is instead converted into a \textsc{P}\(^2\) and consumed directly by any off-the-shelf MLLM, see~\cref{fig:teaser} for illustration.

Our contributions are summarized as follows:
\begin{itemize}
    \item \textbf{Unified Representation.} We introduce Perception Programs (\textsc{P}\(^2\)), a symbolic framework that reformulates visual tool outputs as visually-grounded language-native summary. It generalizes across depth estimation, optical flow, visual and semantic correspondence, jigsaw-style reasoning, and object localization. In addition to requiring no model fine-tuning or multiple tool calls, it works across model scales and architectures.
    \item \textbf{Comprehensive Evaluation.} We show that this representation enables robust reasoning on six BLINK~\citep{fu2024blink} tasks. 

    \textsc{P}\(^2\) yields an average gain of \(19.48\%\) on larger base models (GPT-5 Mini, Gemini 2.5 Pro) relative to raw tool use, while improving smaller open-source models such as InternVL3.5-\(2\)B, InternVL3.5-\(4\)B, and Qwen3VL-\(4\)B by \(22.18\%\). Across all benchmarks, \textsc{P}\(^2\) achieves an average \(19.66\%\) improvement over the prior best results.
    
    \item \textbf{Analysis of Bottlenecks.} 
    We examine GPT-5 Mini's ability to interpret raw visual tool outputs by prompting it to verbalize its understanding in language. For depth maps, the model fails to preserve relative patch ordering—Kendall's $\tau$ rapidly approaches zero as grid resolution increases (see~\cref{fig:gpt5illustration}). For visual correspondence, it often reproduces input tokens rather than reasoning over image content. Moreover, chain-of-thought prompts that encourage explicit verbalization of visual cues produce noisy descriptions and degrade performance.
    We also investigate the effect of augmenting existing agentic tool-use methods with \textsc{P}\(^2\), yielding \(18.28\%\) improvement on depth and localization tasks, demonstrating that \textsc{P}\(^2\) enhances both interpretability and effectiveness in tool-use MLLMs.
\end{itemize}

These results show that when vision tools already provide the necessary visual cues, the primary bottleneck is not additional tool calls or larger models, but the representation of visual tool outputs. \textsc{P}\(^2\) directly addresses this bottleneck by presenting the evidence in a language-native form that MLLMs can reliably parse and reason over.

\section{Related Work}
\label{sec:relatedwork}

\citet{fu2025hiddenplainsightvlms} show that MLLMs often under-use their encoders and lean heavily on language prior, while another work documents limitations in their visual understanding~\citep{tong2024eyes}. Recent efforts attempt to mitigate these limitations either through in-context learning ability of LLMs~\citep{gupta2023visual}, zero-shot prompting~\citep{suris2023vipergpt}, or fine-tuning~\citep{zhang2025thyme} to emit modular programs (in a programming language), designing specialized agent-and-tool collaboration mechanisms~\citep{zhang2024vipact}, or enabling MLLMs to think with images~\citep{ma2025latte,fan2025grit,yang2025machine}.


\paragraph{Tool-Use through Program Synthesis} Here, the MLLM is given an API specification plus a query (and/or image) and is asked to synthesize an executable program whose output is used to answer the question. VisProg~\citep{gupta2023visual} uses in-context examples with GPT-3 to generate Python programs that call vision tools and subroutines. ViperGPT~\citep{suris2023vipergpt} similarly prompts a code-generation model (GPT-3 Codex) with the query and an API specification of available modules (classical image operations, neural models, other LLMs) and executes the resulting Python on the input image or video. Thyme~\citep{zhang2025thyme} fine-tunes Qwen2.5-VL (\(7\)B) in two stages (SFT + RL) so that, for each query, the model can emit code, reasoning, or both before answering. MMFactory~\citep{fan2024mmfactory} maintains a model/tool repository and, given a multimodal query and constraints, proposes programmatic pipelines that compose vision tools and (M)LLMs. While powerful, these systems can be computationally heavy, often require multiple LLM calls and a sand-boxed executor, and may struggle with multi-image settings~\citep{fan2024mmfactory}.

\paragraph{Tool-Use through Chain-of-Thought} An alternative is to integrate tools directly into the reasoning chain instead of emitting full programs. Aurora~\citep{bigverdi2025perception} introduces perception tokens, discrete latent codes that approximate a visual modality, and let the MLLM `call' vision tools within its chain-of-thought. Mirage~\citep{yang2025machine} encourages the model to `visually imagine' by generating latent image representations as intermediate states, and is trained in two stages (latent reinforcement followed by text-only prediction). Visual Sketchpad~\citep{hu2024visual} leverages sketch-like intermediate tool outputs, motivated by human use of sketches for visual reasoning. LATTE~\citep{ma2025latte} builds a multimodal reasoning dataset that includes expert visual tool outputs and fine-tunes a family of MLLMs on it. However, these methods still operate at pixel-level tool outputs and thus inherit under-utilization of vision encoders~\citep{fu2025hiddenplainsightvlms} and limited understanding of dense visual modalities. For further review, see~\cref{appdx:more_rl} (appendix).

\section{Perception Programs}
\label{sec:method}

We formalize a Perception Program (\textsc{P}\(^2\)) as a compact, symbolic summary of sensory inputs, or visual modalities, that standardizes \textit{what} is present, \textit{where} it is located, and optionally \textit{how} different parts relate. \textsc{P}\(^2\) exposes a unified item schema shared across modalities and grounded in language, enabling off-the-shelf LLMs to reason over expert tool outputs through \textit{reading the visual modalities}.

\subsection{General Construction}
Let the pixel domain be \(\Omega = \{0,...,W-1\} \times \{0,...,H-1\}\). We define a finite set of primitives \(\mathcal{P}\), wherein each primitive \(p \in \mathcal{P}\) is associated with a spatial support \(S_{p} \subseteq \Omega\) (e.g., a patch, a point, an image, etc.) and a normalized location \(c_{p} \in \{0,...,1000\}^{2}\). For any pixel coordinate \((x,y)\), we normalize it to a canonical location as \((\lfloor\frac{1000x}{W}\rfloor, \lfloor\frac{1000y}{H}\rfloor)\), and we use this map to define location field where each primitive \(p\) is assigned to a location \(c_{p}\). Note that, when \(S_{p}\) is a region, we take its center and normalize it this way.

For each \(p \in \mathcal{P}\), \textsc{P}\(^2\) emits a structured item
\begin{equation}
    I_{p} = (p, c_{p}, r_{p}, b_{p}),
\end{equation}
where \(p\) is the primitive identifier, \(c_{p}\) is the normalized spatial coordinate, \(r_{p}\) is the reading from the modality data on \(S_{p}\), and \(b_{p}\) is an optional label. \textsc{P}\(^2\) can include a sparse set of symbolic triples \(\mathcal{T}\) denoting relations between the primitives, defined as \((p_a, \pi, p_b)\) wherein \(p_a, p_b \in \mathcal{P}\), and \(\pi\) is the predicate name (e.g., darker than, adjacent to, etc.). These relations are generated by comparing item statistics for candidate pairs. We serialize \textsc{P}\(^2\) as a YAML-like text block summarizing the visual modality in a language-first format. The schema is invariant across modalities, the only changes are: the construction of \(r_{p}\), if \(b_{p}\) is included, and whether relations are emitted. A general algorithm is given in~\cref{alg:pp-construction}. Unless \(\mathcal{T}\) is explicitly mentioned, it should be assumed empty (i.e. \(\mathcal{T} = \emptyset\)).

\subsection{Modality Instantiations} We instantiate the above framework for each visual modality dependent on the task.

\subsubsection{Depth}
An expert depth estimation tool produces a scalar field \(D: \Omega \rightarrow [0,1]\) where larger values indicate proximity (nearer). We then partition this depth field into a regular \(P\times P\) grid. This yields \(K\) disjoint grid cells \(\{S_{1}, S_{2},..., S_{K}\}\) and \(S_{p} \subset \Omega\) indexed in row-major order. Let \(c_{p} \in [0,1000]^{2}\) denote the normalized center of the cell \(S_{p}\), then for each grid cell \(S_{p}\), the \textsc{P}\(^2\) read-out, \(r_{p}\), is computed as
\begin{equation}
\label{eq:depthreadout}
    r_{p} = \left[\min_{(x,y)\in S_{p}}D(x,y), \max_{(x,y)\in S_{p}}D(x,y)\right].
\end{equation}

In other words, for every patch we store the minimum and maximum value observed in that spatial support. For each \((a,b) \in \mathcal{N}\) where \(\mathcal{N}\) is a 4-neighborhood adjacency defined on the same partition \(\{S_{1},...,S_{K}\}\), then depth instantiated \textsc{P}\(^2\) emits relations as
\begin{equation}
\label{eq:depthrelations}
    t_{(a,b)} = 
    \begin{cases}
      (a, \text{in-front of}, b), & \text{if}\ \mu_{a} > \mu_{b}+\tau \\
      (b, \text{in-front of}, a), & \text{if}\ \mu_{b} > \mu_{a}+\tau
    \end{cases},
\end{equation}
where \(\mu_{p} = \frac{1}{|S_{p}|}\sum_{(x,y)\in S_{p}}D(x,y)\) which is the average depth of grid cell \(S_{p}\) and \(\tau > 0\) is a small margin that suppresses relations caused by small differences.

\begin{algorithm}[!t]
\caption{Perception Program (\textsc{P}$^2$) Construction}
\label{alg:pp-construction}
\begin{algorithmic}[1]
\REQUIRE Data $\mathcal{D}$, $(W,H)$
\ENSURE $\textsc{P}^2 = (\mathcal{P}, \{I_p\}, \mathcal{T})$

\STATE $\mathcal{P} \leftarrow$ \texttt{ExtractPrimitives}($\mathcal{D}$)
\FOR{each $p \in \mathcal{P}$}
    \STATE $c_p \leftarrow$ \texttt{ExtractCoordinates}($S_p$)
    \STATE $r_p \leftarrow$ \texttt{ExtractReadOut}($S_p$,\ $\mathcal{D}$)
    \STATE $b_p \leftarrow$ \texttt{ExtractLabel}($p$,\ $\mathcal{D}$)
    \STATE $I_p \leftarrow (p, c_p, r_p, b_p)$
\ENDFOR

\STATE $\mathcal{R} \leftarrow \{r_p \mid p \in \mathcal{P}\}$ \hfill \textcolor{commentcolor}{// collect all primitive read-outs}
\STATE $\mathcal{T} \leftarrow$ \texttt{ExtractRelations}($\mathcal{P}$, $\mathcal{R}$)

\RETURN $\textsc{P}^2 = (\mathcal{P}, \{I_p\}, \mathcal{T})$
\end{algorithmic}
\end{algorithm}

\subsubsection{Optical Flow}
An expert optical flow estimation tool produces a vector field \(F: \Omega \rightarrow \mathbb{R}^{2}\) on the same image domain \(\Omega = \{0,...,W-1\} \times \{0,...,H-1\}\) with \(F(x,y) = (u(x,y), v(x,y))\) where \(u\) and \(v\) denote the horizontal and vertical components, respectively. Without loss of generality, we use horizontal flow. We then partition the field into a regular \(P\times P\) grid. This yields \(K\) disjoint grid cells \(\{S_{1}, S_{2},..., S_{K}\}\) and \(S_{p} \subset \Omega\) indexed in row-major order. Let \(c_{p} \in [0,1000]^{2}\) denote the normalized center of the cell \(S_{p}\), and \(\bar{u}_{p}\) denote the mean horizontal component computed as \(\bar{u}_{p} = \frac{1}{|S_{p}|}\sum_{(x,y)\in S_{p}}u(x,y)\), then for each grid cell \(S_{p}\), the \textsc{P}\(^2\) read-out, \(r_{p}\), is computed as
\begin{equation}
\label{eq:flowreadout}
    r_{p} = 
    \begin{cases}
        \text{`left'}, & \text{if}\ \bar{u}_{p} < 0, \\
        \text{`right'}, & \text{if}\ \bar{u}_{p} \geq 0.
    \end{cases}
\end{equation}

\subsubsection{Visual Correspondence}
Visual correspondence problems have two views of the same scene with some varying condition (e.g. light or camera position) and the task is to find matching points between them. A correspondence tool produces multiple point-to-point matches between a reference image and a target image. Let the reference image have a spatial domain \(\Omega_{1} = \{0,...,W_{1}-1\} \times \{0,...,H_{1}-1\}\), and the target image have domain \(\Omega_{2} = \{0,...,W_{2}-1\} \times \{0,...,H_{2}-1\}\). The tool outputs a set of \(N\) correspondences \(\{((x_{i}^{(1)}, y_{i}^{(1)}), (x_{i}^{(2)},y_{i}^{(2)}))\}^{N}_{i=1}\), where \((x_{i}^{(1)}, y_{i}^{(1)}) \in \Omega_{1}\) is a keypoint in the reference image and \((x_{i}^{(2)}, y_{i}^{(2)}) \in \Omega_{2}\) is a matched keypoint in the target image. We do not form a grid in this modality. For each match we normalize both the reference and target locations to \([0,1000]^{2}\), then set the former as \(c_{i}\) and the latter as \(r_{i}\).

\subsubsection{Jigsaw}

The jigsaw problem involves an image with a missing piece and a set of candidate images that may complete it. In our setting, the missing region corresponds to the lower-right corner of the source image, and we are given two candidate pieces, denoted \(i \in \{A, B\}\). Each primitive in this configuration represents a combination between a candidate piece and one of its relevant edges, where edges are defined with respect to the candidate image itself (hence the names \textit{left} and \textit{top}). Thus, the set of primitives is
\begin{equation}
    \mathcal{P} = \{(\text{left}, A), (\text{top}, A), (\text{left}, B), (\text{top}, B)\}.
\end{equation}

For each primitive \(p = (b,i)\), the coordinates \(c_p = (x_0, y_0, x_1, y_1)\) denote the location of the edge in the coordinate space of candidate \(i\), normalized to the range \([0,1000]^4\). The read-out \(r_p\) is defined as the average of structural, edge, and color similarity scores between the border strip close to the missing region in the reference image and the corresponding border strip of the candidate, and \(r_p \in [0,1]\).

\begin{figure*}
    \centering
    \includegraphics[width=\linewidth]{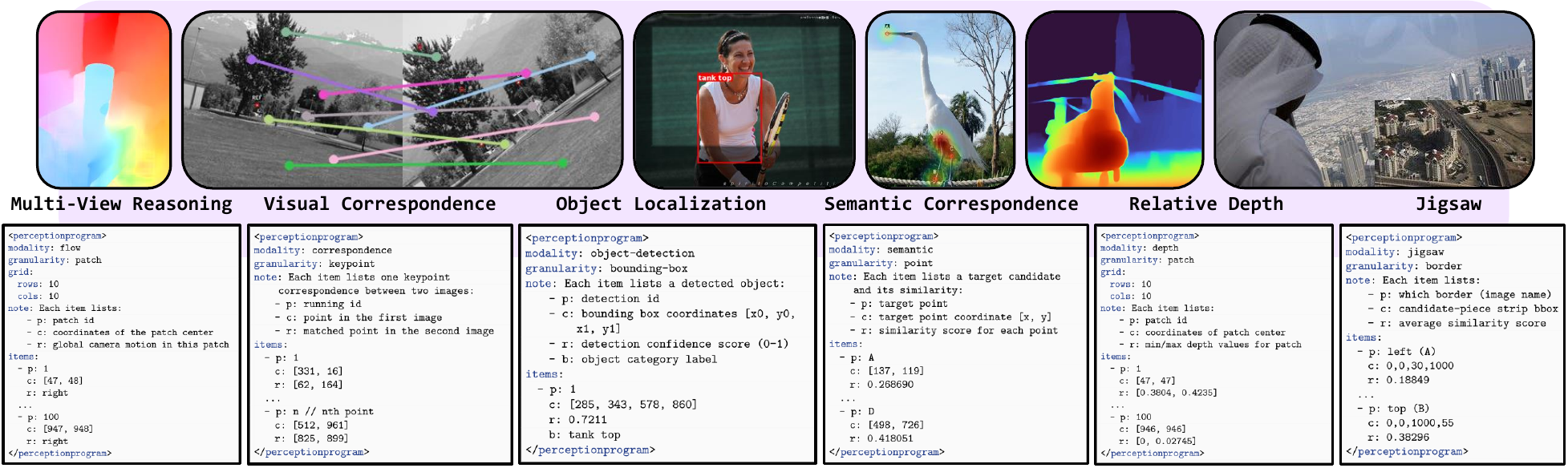}
    \caption{\textbf{Perception Program Instantiations.} Top: Tool outputs. Bottom: \textsc{P}\(^2\) instantiations of those respective tools.}
    \label{fig:sample_pps_with_tools}
\end{figure*}

\subsubsection{Object Detection}
An object detection tool provides a set of detections for an input image with each detection consisting of category label, confidence score \(\in [0,1]\), and a bounding box localizing the object. We take the output from the detector as-is and treat each detection as one primitive \(p \in \mathcal{P}\). Given an image size \((W,H)\), we normalize the coordinates to the common \([0,1000]^{2}\) range, following similar procedure as described earlier, and, with slight abuse of notation, set this at the location field in \textsc{P}\(^2\) for that primitive, i.e., \(c_{p} = (x_{0}, y_{0}, x_{1}, y_{1})\). The read-out is the confidence score, while \(b_{p}\) is the label of the object (category).

\subsubsection{Semantic Correspondence}
The semantic correspondence problem seeks to identify matching points between two related images based on their visual or geometric similarity. An expert tool suited for this task is a feature extractor, which takes a pair of points, one from each image, and computes a similarity score between their corresponding features. In our case, we consider one point in the source image and four candidate points in the target image, denoted \(A\), \(B\), \(C\), and \(D\). Accordingly, we define the primitive set as,
\begin{equation}
    \mathcal{P} = \{A, B, C, D\}.
\end{equation}

Given the discrete nature of this modality, we do not form a grid. The coordinate \(c_p\) of each primitive represents the normalized pixel location of the corresponding candidate point in the target image, and the read-out \(r_p\) is the similarity score produced by the expert tool. We do not include an optional label \(b_p\) since the candidate identifier itself serves that role.

\section{Evaluation Setup}
\label{sec:exps}

We posit that Perception Program (\textsc{P}\(^2\)) lets MLLMs \textit{read visual modalities}. To test this, we consider BLINK benchmark~\citep{fu2024blink}, a suite of \(14\) perception-focused tasks. We concentrate evaluation on six sub-tasks from the benchmark where additional modalities are especially informative. We benchmark multiple reasoning MLLMs with and without \textsc{P}\(^2\), and for each task, specify the modality-aware tool used to facilitate \textsc{P}\(^2\) construction.

\subsection{Tasks \& Tools}
\label{subsec:tasksandtools}
We evaluate six BLINK sub-tasks, namely multi-view reasoning, relative depth, visual correspondence, jigsaw, semantic correspondence, and object localization. We exclude datasets centered on IQ testing or commonsense compositional reasoning, as they do not directly assess visual perception. Instead of BLINK's standard relative depth task, which considers pairwise point comparisons, we adopt HardBLINK, the harder variant introduced in~\citet{bigverdi2025perception}. The study introduces three increasingly challenging settings with three, four, and five comparative points, respectively. 

Each task is paired with an off-the-shelf modality/tool that instantiates the \textsc{P}\(^2\). Concretely, HardBLINK (relative depth) uses estimated monocular depth from DepthAnything~\citep{yang2024depth}, multi-view reasoning uses optical flow estimated with RAFT~\citep{teed2020raft}, visual correspondence uses dense feature matching with LoFTR~\citep{sun2021loftr}, jigsaw uses a mix of Structural Similarity Index Measure (SSIM)~\citep{wang2004image}, HSV-hist $\chi^2$ and gradient-based normalized cross-correlation (NCC) for structural alignment, color compatibility and edge continuity respectively, semantic correspondence uses feature similarity computed with DIFT~\citep{tang2023emergent}, and object localization uses open-vocabulary object detection computed with LLMDet~\citep{fu2025llmdet}. BLINK tasks have multiple alternative options (points) overlaid on the images. We provide their coordinates to the models as an additional \textsc{P}\(^2\).

\begin{table*}[!t]
\centering
\scalebox{0.8}{
\begin{tabular}{@{}clcccccc@{}}
\thickhline
\multicolumn{2}{c}{} & \multicolumn{1}{c}{\begin{tabular}[c]{@{}c@{}}Multi-View\\ Reasoning\end{tabular}} & \multicolumn{1}{c}{\begin{tabular}[c]{@{}c@{}}HardBLINK\\ Depth\end{tabular}} & \multicolumn{1}{c}{\begin{tabular}[c]{@{}c@{}}Visual\\ Correspondence\end{tabular}} & \multicolumn{1}{c}{Jigsaw} & \multicolumn{1}{c}{\begin{tabular}[c]{@{}c@{}}Object\\ Localization\end{tabular}} & \multicolumn{1}{c}{\begin{tabular}[c]{@{}c@{}}Semantic\\ Correspondence\end{tabular}} \\ \midrule
\multicolumn{2}{c}{Human~\citep{fu2024blink}} & \(92.48\) & N/A & \(99.42\) & \(99.00\) & \(98.00\) & \(96.07\) \\
\multicolumn{2}{c}{\fire~Thyme~\citep{zhang2025thyme}} & \(51.88\) & \(35.21\) & \(63.95\) & \(58.00\) & \(48.36\) & \(39.57\) \\
\multicolumn{2}{c}{\fire~LATTE~\citep{ma2025latte}} & \(48.87\) & N/A & \(29.65\) & \(75.33\) & \(47.54\) & \(33.09\) \\
\multicolumn{2}{c}{\snow~Visual Sketchpad (GPT-4o)~\citep{hu2024visual}} & \(45.60\) & N/A & \(80.80\) & \(70.70\) & \(65.40\) & \(58.30\) \\ 
\multicolumn{2}{c}{\snow~Visual Sketchpad (GPT-5 Mini)\(^\star\)~\citep{hu2024visual}} & \(47.37\) & \(63.71\) & \(91.86\) & \(79.33\) & \(67.21\) & \(60.43\) \\
\multicolumn{2}{c}{\snow~MMFactory (GPT-4o)~\citep{fan2024mmfactory}} & \(60.20\) & N/A & \(85.50\) & \(75.30\) & \(59.00\) & \(58.30\) \\
\multicolumn{2}{c}{\fire~TULIP (Llama-3.2-\(11\)B)~\citep{tang2025tulip}} & \(44.96\) & N/A & \(48.97\) & \(57.26\) & \(60.01\) & \(29.61\) \\
\multicolumn{2}{c}{\fire~PerceptionLM-\(8\)B~\citep{cho2025perceptionlm}} & \(55.64\) & \(51.08\) & \(43.60\) & \(40.00\) & \(63.11\) & \(38.13\) \\
\multicolumn{2}{c}{\fire~Zebra-CoT (Anole \(7\)B)~\citep{li2025zebra}} & \(21.02\) & N/A & \(26.16\) & \(39.33\) & \(45.90\) & \(17.99\) \\
\multicolumn{2}{c}{\fire~OVR-\(7\)B~\citep{wei2025open}} & \(46.62\) & \(25.27\) & \(32.56\) & \(53.33\) & \(51.64\) & \(24.50\) \\
\multicolumn{2}{c}{Prior State-of-the-Art} & \(60.20\)~(\snow)~\citep{fan2024mmfactory} & \(61.56\)~(\fire)~\citep{zhou2025reinforced} & \(85.50\)~(\snow)~\citep{fan2024mmfactory} & \(88.00\)~(\fire)~\citep{yang2025machine} & \(65.40\)~(\snow)~\citep{hu2024visual} & \(58.30\)~(\snow)~\citep{hu2024visual,fan2024mmfactory} \\ 
\midrule
\multirow{3}{*}{\begin{tabular}[c]{@{}c@{}}\snow~GPT-5 Mini~\citep{gpt5team}\\ (Zero-Shot) \end{tabular}} & Standard & \(41.35\) & \(52.42\) & \(76.74\) & \(76.00\) & \(58.20\) & \(53.24\) \\
 & Raw Tool & \(45.11\) & \(65.05\) & \(75.58\) & \(66.00\) & \(59.02\) & \(53.24\) \\
 & \cellcolor{rowcolor} \textsc{P}\(^2\) (Ours) & \cellcolor{rowcolor} \(\mathbf{86.47}\) & \cellcolor{rowcolor} \(\mathbf{81.45}\) & \cellcolor{rowcolor} \(\mathbf{94.19}\) & \cellcolor{rowcolor} \(\mathbf{91.33}\) & \cellcolor{rowcolor} \(\mathbf{93.44}\) & \cellcolor{rowcolor} \(\mathbf{64.03}\) \\ \midrule
\multirow{3}{*}{\begin{tabular}[c]{@{}c@{}}\snow~Gemini 2.5 Pro~\citep{comanici2025gemini}\\ (Zero-Shot) \end{tabular}} & Standard &  \(48.12\) & \(57.79\) & \(91.86\) & \(78.00\) & \(70.49\) & \(64.03\) \\
 & Raw Tool & \(33.83\) & \(61.83\) & \(91.86\) & \(88.00\) & \(57.38\) & \(66.19\) \\
 & \cellcolor{rowcolor} \textsc{P}\(^2\) (Ours) & \cellcolor{rowcolor} \(\mathbf{63.91}\) & \cellcolor{rowcolor} \(\mathbf{77.68}\) & \cellcolor{rowcolor} \(\mathbf{96.51}\) & \cellcolor{rowcolor} \(\mathbf{96.67}\) & \cellcolor{rowcolor} \(\mathbf{77.05}\) & \cellcolor{rowcolor} \(\mathbf{74.10}\) \\ \midrule
\multirow{3}{*}{\begin{tabular}[c]{@{}c@{}}\snow~Qwen3VL-\(4\)B~\citep{qwen3technicalreport}\\ (One-Shot)\end{tabular}} & Instruct &  \(39.84\) & \(41.66\) & \(80.23\) & \(79.33\) & \(54.10\) & \(59.71\)  \\
 & Standard & \(45.90\) & \(47.04\) & \(86.62\) & \(64.67\) & \(54.92\) & \(61.87\) \\
 & Raw Tool & \(67.66\) & \(42.40\) & \(76.16\) & \(60.66\) & \(45.90\) & \(62.58\) \\
 & \cellcolor{rowcolor} \textsc{P}\(^2\) (Ours) & \cellcolor{rowcolor} \(\mathbf{93.98}\) & \cellcolor{rowcolor} \(\mathbf{61.02}\) & \cellcolor{rowcolor} \(\mathbf{88.37}\) & \cellcolor{rowcolor} \(\mathbf{86.67}\) & \cellcolor{rowcolor} \(\mathbf{85.25}\) & \cellcolor{rowcolor} \(\mathbf{64.03}\) \\ \midrule
\multirow{3}{*}{\begin{tabular}[c]{@{}c@{}}\snow~InternVL3.5-\(2\)B~\citep{wang2025internvl3}\\ (One-Shot)\end{tabular}} & Standard &  \(49.62\) & \(29.57\) & \(34.88\) & \(61.33\) & \(47.54\) & \(33.81\) \\
 & Raw Tool & \(56.39\) & \(25.00\) & \(\mathbf{41.86}\) & \(62.00\) & \(50.00\) & \(34.53\) \\
 & \cellcolor{rowcolor} \textsc{P}\(^2\) (Ours) & \cellcolor{rowcolor} \(\mathbf{69.17}\) & \cellcolor{rowcolor} \(\mathbf{32.25}\) & \cellcolor{rowcolor} \(39.53\) & \cellcolor{rowcolor} \(\mathbf{83.33}\) & \cellcolor{rowcolor} \(\mathbf{55.74}\) &\cellcolor{rowcolor} \(\mathbf{46.04}\) \\ \midrule
\multirow{3}{*}{\begin{tabular}[c]{@{}c@{}}\snow~InternVL3.5-\(4\)B~\citep{wang2025internvl3}\\ (One-Shot)\end{tabular}} & Standard & \(45.86\) & \(40.59\) & \(65.69\) & \(80.00\) & \(56.56\) & \(45.32\) \\
 & Raw Tool & \(75.18\) & \(37.90\) & \(64.53\) & \(66.67\) & \(54.92\) & \(44.60\) \\
 & \cellcolor{rowcolor} \textsc{P}\(^2\) (Ours) & \cellcolor{rowcolor} \(\mathbf{94.73}\) & \cellcolor{rowcolor} \(\mathbf{69.89}\) & \cellcolor{rowcolor} \(\mathbf{86.04}\) & \cellcolor{rowcolor} \(\mathbf{90.67}\) & \cellcolor{rowcolor} \(\mathbf{90.16}\) & \cellcolor{rowcolor} \(\mathbf{62.59}\) \\
\thickhline
\end{tabular}
}
\caption{\textbf{Results on BLINK~\citep{fu2024blink}.} Accuracy (\(\%\)) on six perception-centric sub-tasks. We report three settings per model, Standard (image+question only), Raw Tool (tool output as auxiliary input), and \colorbox{rowcolor}{\textsc{P}\(^2\)} (our training-free Perception Program instantiated from same tool output). Further, we also list representative prior tool-use or perception-oriented methods, plus previous state-of-the-art for each sub-task. Bold indicates best score within each model block. Even with the much smaller InternVL3.5-\(4\)B and Qwen3VL-\(4\)B MLLMs than prior work, \colorbox{rowcolor}{\textsc{P}\(^2\)} yields large gains setting new state-of-the-art results. \(^\star\) indicates we ran with GPT-5 Mini as LLM and official codebase.~\fire~denotes training/fine-tuning, and~\snow~indicates no parameter updates necessary.}
\label{tab:mainresults}
\end{table*}

\subsection{MLLMs Evaluated} 
We evaluate a representative mix of frontier and open-source/open-weight MLLMs: GPT-5 Mini (2025-08-07)~\citep{gpt5team}, Gemini 2.5 Pro~\citep{comanici2025gemini}, InternVL3.5-\(2\)B and \(4\)B~\citep{wang2025internvl3}, and Qwen3VL-\(4\)B~\citep{qwen3technicalreport}, and report accuracy on the validation split of BLINK following the original work~\citep{fu2024blink}. For each of the MLLM, we use its reasoning variant and evaluate three settings: (i) Standard, wherein the model is queried as is with only the images and question, (ii) Tool, where the model additionally receives the raw tool output as an auxiliary input (simulating a tool-calling pipeline), and (iii) \textsc{P}\(^2\), wherein the raw tool output is first converted into a Perception Program and then is supplied, alongside the images and question, to the model.

Notably, tools are as outlined in~\cref{subsec:tasksandtools}, the main difference between settings (ii) and (iii) being that the former provides the information directly as pixels, whereas the latter processes and digests it into the \textsc{P}\(^2\) template. For jigsaw, we follow prior work~\citep{yang2025machine,hu2024visual} and choose the raw tool to be the two images corresponding to trial-and-error of the two candidates (i.e. tentative images with bottom right corners overlaid with the candidate completions), causing the MLLM's task to be simply the recognition of global image consistency, see~\cref{fig:sample_pps_with_tools} for tools and their \textsc{P}\(^2\) instantiations.

\subsection{Baselines} 
We also benchmark against five strands of prior work aimed at strengthening perceptual understanding in MLLMs: (i) chain-of-thought methods that reason over tool outputs, e.g., LATTE~\citep{ma2025latte}, Thyme~\citep{zhang2025thyme}, Aurora~\citep{bigverdi2025perception}, and Mirage~\citep{yang2025machine}; (ii) architectures with dedicated perception-oriented modules, e.g., TULIP~\citep{tang2025tulip} and PerceptionLM~\citep{cho2025perceptionlm}; (iii) data-centric pipelines that improve supervision and training signals, e.g., Zebra-CoT~\citep{li2025zebra}; (iv) agentic frameworks that leverage tools during inference, e.g., Visual Sketchpad~\citep{hu2024visual}, MMFactory~\citep{fan2024mmfactory}; and (v) reinforcement-learning based reasoning pipelines, e.g., ReVPT~\citep{zhou2025reinforced}, ViGoRL~\citep{sarch2025grounded}, and OVR~\citep{wei2025open}.

\section{Results}
\label{sec:results}

\begin{figure}
    \centering
    \includegraphics[width=0.9\linewidth]{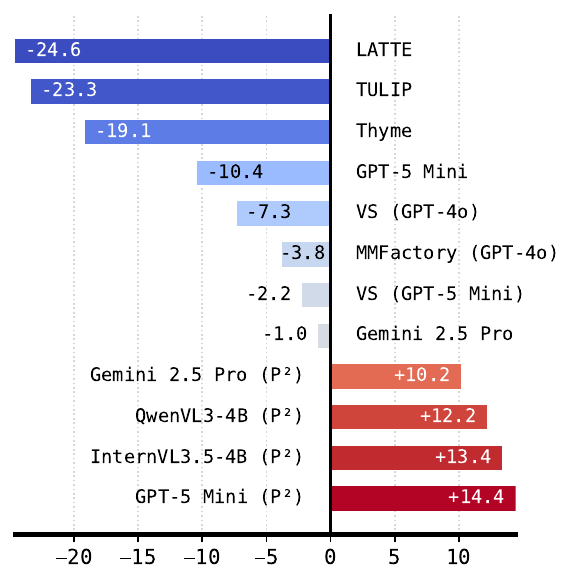}
    \caption{\textbf{Mean \(\Delta\) vs. prior SOTA across BLINK.} Bars show average accuracy improvement (percentage points) of each method over task-wise (except HardBLINK) prior state-of-the-art (at point zero; see~\cref{tab:mainresults}). Positive values indicate gains over prior SOTA; negative values indicate regressions. Numeric \(\Delta\) are written inside/beyond the bars along with their method names. VS denotes Visual Sketchpad.}
    \label{fig:results_plot}
\end{figure}

\Cref{tab:mainresults,tab:more_results_1} (summary in~\cref{fig:results_plot}) summarize performance across six BLINK~\citep{fu2024blink} sub-tasks. Across all models and tasks, converting raw tool outputs into language-native Perception Program (\textsc{P}\(^2\)) consistently and often substantially outperforms both Standard (image+question only) and Raw Tool (raw tool as auxiliary input) settings. The gap between Raw Tool and \textsc{P}\(^2\) highlights that merely appending tool outputs can be neutral or even harmful (e.g., Gemini 2.5 Pro on multi-view reasoning reduces by \(14.29\%\)).

On stronger models, \textsc{P}\(^2\) sets new state-of-the-art results on every task we consider. With GPT-5 Mini, \textsc{P}\(^2\) surpasses prior best results by a large margin, including performing better than Visual Sketchpad~\citep{hu2024visual} with GPT-5 Mini backbone. With the same GPT-5 Mini backbone, \textsc{P}\(^2\) also uses significantly less average tokens per sample than Visual Sketchpad~\citep{hu2024visual}, see~\cref{fig:tokusage} (in appendix). Further, Gemini 2.5 Pro shows similar trends, often making even better use of \textsc{P}\(^2\), e.g., on semantic correspondence, despite the standard Gemini 2.5 Pro already performing similarly to \textsc{P}\(^2\)-equipped GPT-5 Mini, its \textsc{P}\(^2\) performance improves by \(10\%\) rather than saturating. Smaller models benefit as well. InternVL3.5-\(2\)B~\citep{wang2025internvl3} sees sturdy improvements from Standard to \textsc{P}\(^2\). While InternVL3.5-\(2\)B is limited in general, its \(4\)B variant matches performance of base GPT-5 Mini and Gemini 2.5 Pro when coupled with \textsc{P}\(^2\). Qwen3VL-\(4\)B (reasoning variant) observes gains markedly under \textsc{P}\(^2\).

\begin{table}[!t]
    \centering
    \resizebox{0.9\linewidth}{!}{%
    \begin{tabular}{@{}lccc@{}}
    \thickhline
    \multicolumn{1}{c}{\textbf{Methods}} & \begin{tabular}[c]{@{}c@{}}\textbf{Visual}\\ \textbf{Correspondence}\end{tabular} & \textbf{Jigsaw} & \begin{tabular}[c]{@{}c@{}}\textbf{Semantic}\\ \textbf{Correspondence}\end{tabular} \\ \midrule
    \fire~Mirage (Qwen2.5VL-\(7\)B)~\citep{yang2025machine} & N/A & \(88.00\) & N/A \\
    \fire~Mirage (Qwen2.5VL-\(3\)B)~\citep{yang2025machine} & N/A & \(85.00\) & N/A \\
    \fire~\begin{tabular}[c]{@{}l@{}}\citet{zhang2025improving} (Qwen2.5VL-\(7\)B)\\ \end{tabular} & \(69.77\) & \(74.67\) & N/A \\
    \fire~ViGoRL~\citep{sarch2025grounded} & N/A & \(56.00\) & N/A \\
    \snow~GPT-4 Turbo & \(48.80\) & \(64.70\) & \(30.90\) \\
    \snow~\begin{tabular}[c]{@{}l@{}}Visual Sketchpad~\citep{hu2024visual} (GPT-4 Turbo)\\ \end{tabular} & \(52.30\) & \(68.50\) & \(42.40\) \\ 
    \snow~LLaVA-\(13\)B~\citep{liu2024llava} & \(29.10\) & \(58.0\) & \(32.40\) \\
    \snow~\begin{tabular}[c]{@{}l@{}}MMFactory (LLaVA-\(13\)B)~\citep{fan2024mmfactory}\\ \end{tabular} & \(34.30\) & \(64.00\) & \(34.50\) \\
    \midrule
    \snow~InternVL3.5-\(2\)B & \(34.88\) & \(61.33\) & \(33.81\) \\
    \snow~InternVL3.5-\(2\)B+\textsc{P}\(^2\) & \cellcolor{rowcolor} \(\mathbf{39.53}\) & \cellcolor{rowcolor} \(\mathbf{83.33}\) & \cellcolor{rowcolor} \(\mathbf{46.04}\) \\ \midrule
    \snow~InternVL3.5-\(4\)B & \(65.69\) & \(80.00\) & \(45.32\) \\
    \snow~InternVL3.5-\(4\)B+\textsc{P}\(^2\) & \cellcolor{rowcolor} \(\mathbf{86.04}\) & \cellcolor{rowcolor} \(\mathbf{90.67}\) & \cellcolor{rowcolor} \(\mathbf{62.59}\) \\ \midrule
    \snow~Qwen3VL-\(4\)B & \(86.62\) & \(64.67\) & \(61.87\) \\ 
    \snow~Qwen3VL-\(4\)B+\textsc{P}\(^2\) & \cellcolor{rowcolor} \(\mathbf{88.37}\) & \cellcolor{rowcolor} \(\mathbf{86.67}\) & \cellcolor{rowcolor} \(\mathbf{64.03}\) \\ 
    \thickhline
    \end{tabular}}
    \caption{\textbf{Additional BLINK Results.} Comparison of task-specific methods, i.e., methods that do not report on entire BLINK benchmark, on three sub-tasks.~\fire~denotes training/fine-tuning, and~\snow~indicates no parameter updates necessary.}
    \label{tab:more_results_1}
\end{table}

We also compare \textsc{P}\(^2\) to several methods that reason over tool outputs. For multi-view reasoning, visual and semantic correspondence tasks, the strongest prior tool method is MMFactory~\citep{fan2024mmfactory} (\(60.20, 85.50, 58.30\), respectively). Even though it uses GPT-4o as the LLM, both InternVL3.5-\(4\)B and Qwen3VL-\(4\)B outperform it when coupled with \textsc{P}\(^2\). Crucially, several baselines that reason over tools outputs (Thyme~\citep{zhang2025thyme}, LATTE~\citep{ma2025latte}) still trail the much smaller InternVL3.5-\(4\)B and Qwen3VL-\(4\)B coupled with \textsc{P}\(^2\), see~\cref{fig:results_plot}. Overall, our results validate our premise that \textsc{P}\(^2\) turns the same tool signal into a representation that MLLMs can actually \textit{read}, delivering consistent, large, and architecture-agnostic gains without any training or modifications to the underlying MLLM.

\subsection{Analysis}
\label{subsec:analysis}
We first study the quality of visual interpretation of current MLLMs and then investigate the performance gain of plugging \textsc{P}\(^2\) into existing frameworks.
\subsubsection{Quality of Visual Interpretation}
The main goal of this study is twofold: to obtain a more holistic picture of how much information is lost when tasking the MLLM with fine-grained visual interpretation, and to see whether latent reasoning abilities emerge when the MLLM is directed to spell out its own interpretation of the image. For the former goal, we note that, while allowing for efficient testing of the model's ability, the coarse granularity of multiple-choice question answering leads to less informative insights. For the latter, we take inspiration from chain-of-thought prompting and study whether that extends to finer-grained vision tasks.

\begin{figure}
    \centering
    \includegraphics[width=0.9\linewidth]{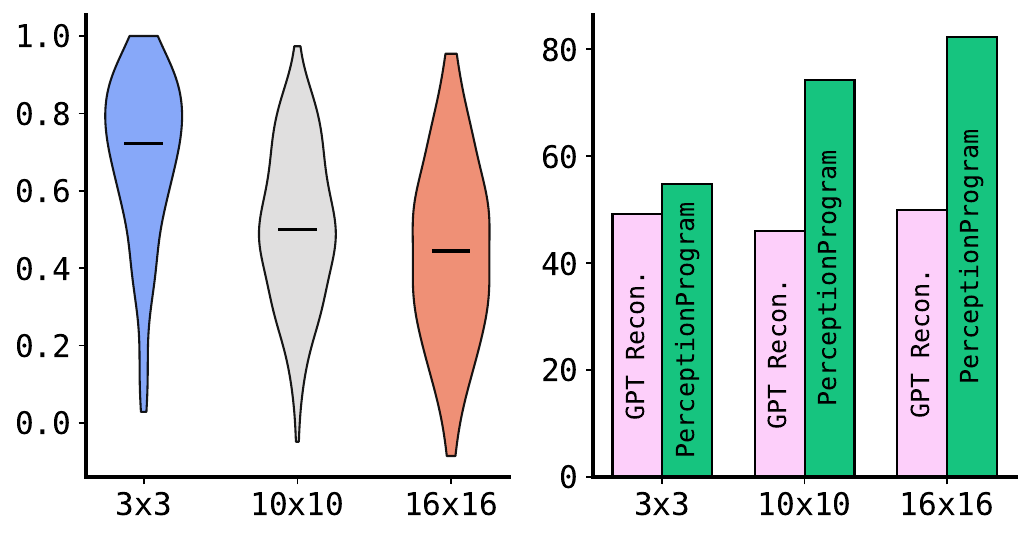}
    \caption{\textbf{GPT-5 Depth Modality Analysis.} Left: Kendall's Tau (y-axis) between ground-truth and GPT-5 Mini reconstructed \textsc{P}\(^2\) decreases as the grid is refined (x-axis). Right: HardBLINK-\(5\) accuracy (y-axis) using GPT-5 Mini's reconstructions (GPT Recon.) across grids (x-axis).}
    \label{fig:depth_recon_ablation}
\end{figure}

\textbf{Relative Depth.} We provide GPT-5 Mini with a set of instructions followed by five image-\textsc{P}\(^2\) pairs to exemplify the conversion, then provide an incomplete \textsc{P}\(^2\) with the depth range redacted and ask the model to write it. \Cref{fig:gpt5illustration} illustrates one such reconstruction. The left panel of~\cref{fig:depth_recon_ablation} illustrates how the relative ordering (i.e. the model's notion of what is the closest, second closest, etc.) diminishes with higher grid size. For each granularity, we plot the distribution of per-sample Kendall \(\tau\) between the ground-truth depth map and its reconstruction on HardBLINK. This quantity illustrates the ranking agreement between the two variants: values close to 1 mean perfect agreement, whereas values close to zero imply the grid orderings are uncorrelated. We see that the reconstruction has reasonable values for the $3 \times 3$ grid, but its information gets quickly corrupted on finer grids.

The right panel shows evaluation of these reconstructed \textsc{P}\(^2\) on HardBLINK-\(5\)~\citep{bigverdi2025perception}, compared to using original \textsc{P}\(^2\). The reconstructed \textsc{P}\(^2\) fails to be useful: GPT-based performance remains roughly flat around \(50\%\) across all grid sizes, whereas \textsc{P}\(^2\) accuracy improves with finer grids, creating a large gap at \(16\times 16\).

\textbf{Visual Correspondence.} Given two images with corresponding points, we study how well the model can follow a line connecting them. We provide the original image pair, and a variant with lines connecting the points (see~\cref{fig:sample_pps_with_tools}), along with a few examples and partially redacted \textsc{P}\(^2\) with `r' fields (i.e. final image coordinates) masked out. The LLM task is to fill them. Problem difficulty is increased with higher displacement between matching points, otherwise a model that neglects visual interpretation and simply copies `\(c\)' field (i.e. initial image coordinates) can attain reasonable reconstructions.

In \Cref{fig:vcorr_recon_ablation}, the left panel plots reconstruction error versus true displacement. Each hexagon aggregates matches, with color indicating its count. Bins forming a narrow band near the x-axis would be indicative of good reconstruction performance. Instead, the errors are large and form a pronounced diagonal structure. Points along it indicate that the MLLM simply copied the input it was given\footnote{We note that the BLINK visual correspondence has a prevalence of images with low camera movement, see supplementary material for more details}. In the right panel, we compare the performance of GT and reconstructed \textsc{P}\(^2\) on the BLINK task. We additionally contrast them with two oracles: a naive algorithm that simply considers Euclidean distance between the reference and alternatives and an oracle procedure that follows the correspondences perfectly (see appendix). In accordance to our hypothesis, the reconstructed performance is worse than the naive baseline, whereas the correct \textsc{P}\(^2\) makes performance competitive with the oracle.

\subsubsection{Plug-and-Play Perception Program}

In \cref{tab:ablation_vs_pp}, we study the performance gain Visual Sketchpad observes if we replace some tools with their \textsc{P}\(^2\) variants. In particular, we post-process outputs of DepthAnything~\citep{yang2024depth} and GroundingDINO~\citep{liu2024grounding}, in addition to providing a tool that gives the coordinates of points A, B, C, D, E upon request. The table shows steady improvement on both tasks, with GPT-5 Mini as the base LLM. Interestingly, the way each model reasons with \textsc{P}\(^2\) output is qualitatively diverse: Visual Sketchpad writes and executes code to process \textsc{P}\(^2\), whereas our variants from \cref{tab:mainresults} interpret it directly.

\subsection{Design Discussions \& Limitations}
\label{subsec:limits}

While \textsc{P}\(^2\) enables MLLMs to \textit{read} visual modalities, here, we also discuss the limitations thereof. 

\textbf{Scope.} Notably, we evaluate six BLINK sub-tasks that have clear tool surrogates. Broader settings (e.g., \(3\)D reasoning beyond depth, general VQA, etc.) may require richer or hierarchical programs, which we do not study here. We deliberately scope this work to perception tasks where tools directly benefit because this isolates the contribution of \textsc{P}\(^2\) as a representation, enabling fair understanding of whether formatting the same evidence lets MLLMs use it effectively.

\textbf{Tool Pairings.} We adopt tool pairings suggested by the original work BLINK~\citep{fu2024blink}, and we note many of our tool choices overlap with tools chosen in LATTE~\citep{ma2025latte} and Visual Sketchpad~\citep{hu2024visual}. However, unlike LATTE's and Visual Sketchpad's agentic approach to tool selection, our setting does not explore tool sequencing, or composition at inference time. This is merely a design decision, and we reiterate that \textsc{P}\(^2\) is trivially pluggable into any agent-and-tool pipeline (e.g., feeding each tool's output through its instantiated \textsc{P}\(^2\)). We leave dynamic tool selection and composition to future. 

\textbf{Tool Reliability.} As with prior methods, \textsc{P}\(^2\) conveys whatever evidence the upstream tool produces, i.e., errors propagate into the program. Our method does not attempt to calibrate or reconcile conflicting tools. Nevertheless, we observe that frontier LLMs, such as GPT-5 Mini and Gemini 2.5 Pro, often utilize \textsc{P}\(^2\) to narrow down choices (in multiple choice questions) and cross-check evidence from \textsc{P}\(^2\) as part of their reasoning before generating the final answer.

\begin{figure}
    \centering
    \includegraphics[width=\linewidth]{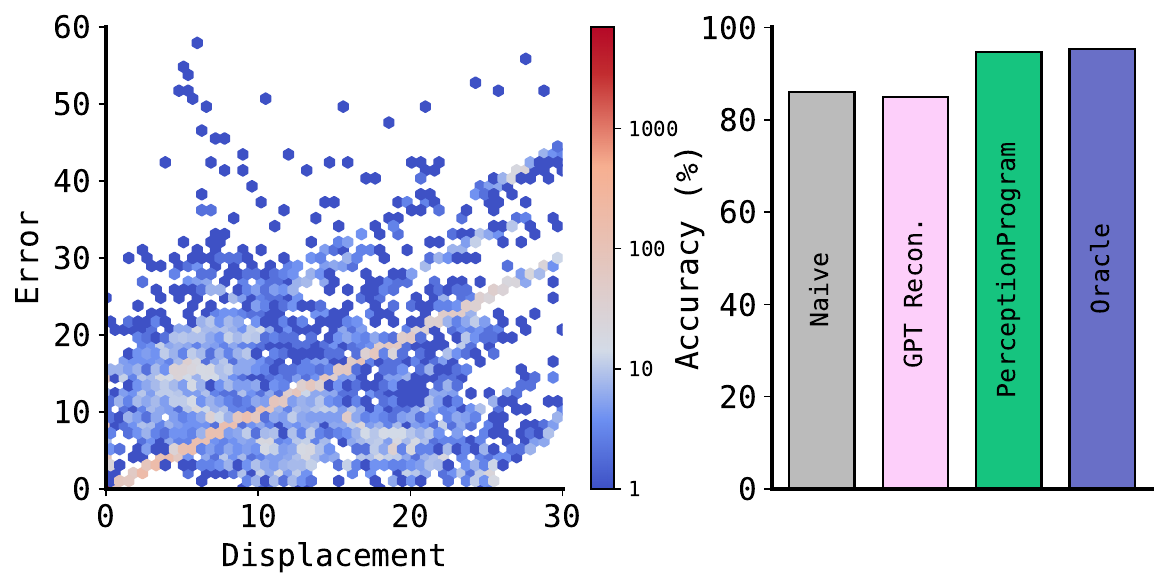}
    \caption{\textbf{GPT-5 Visual Correspondence Modality Analysis.} Left: Plot of LoFTR~\citep{sun2021loftr} correspondences, with true displacement between views on the x-axis and GPT 5’s reconstruction error on the y-axis (both as \% of image diagonal). Color indicates bin count. The strong diagonal indicates a failure mode where GPT often copies the left-image coordinate into the right-image field. Right: Accuracy (\%) across \(100\) matches. Naive refers to Euclidean distance baseline, while Oracle uses the tool directly.}
    \label{fig:vcorr_recon_ablation}
\end{figure}

\begin{table}[!t]
\centering
\resizebox{0.9\linewidth}{!}{
\begin{tabular}{@{}lcccc@{}}
\thickhline
\multicolumn{1}{c}{\multirow{2}{*}{Methods}} & \multicolumn{3}{c}{\textbf{HardBLINK}} & \multirow{2}{*}{\begin{tabular}[c]{@{}c@{}}Object\\ Localization\end{tabular}} \\ \cmidrule(lr){2-4}
\multicolumn{1}{c}{} & \(\mathbf{3}\) & \(\mathbf{4}\) & \(\mathbf{5}\) &  \\ \midrule
\snow~VS~\citep{hu2024visual} & \(71.77\) & \(62.90\) & \(56.45\) & \(60.43\)  \\
\snow~VS + \textsc{P}\(^2\) & \cellcolor{rowcolor} \(\mathbf{81.45}\) & \cellcolor{rowcolor} \(\mathbf{83.06}\) & \cellcolor{rowcolor} \(\mathbf{79.84}\) & \cellcolor{rowcolor} \(\mathbf{80.33}\) \\ 
\thickhline
\end{tabular}}
\caption{\textbf{Plug-and-Play.} Visual Sketchpad (VS) results with \textsc{P}\(^2\).}
\label{tab:ablation_vs_pp}
\end{table}

\section{Conclusion}
\label{sec:conc}

In this work, we proposed Perception Programs (\textsc{P}\(^2\)) that re-write dense tool outputs into compact, symbolic, language-native summaries that specify what cue is present and where it is grounded. We show that this training-free, model-agnostic interface consistently improves performance on several perception-centric tasks, across both frontier and open-source MLLMs, outperforming prior agentic, and tool-use methods. We view \textsc{P}\(^2\) as a step toward a more principled interface between perception and language, suggesting that better \emph{representations} of existing tool signals can be as important as, if not more than, additional tools or larger models.

{
    \small
    \bibliographystyle{ieeenat_fullname}
    \bibliography{main}
}

\clearpage
\setcounter{page}{1}
\maketitlesupplementary

\begin{table}[!t]
    \centering
    \resizebox{0.9\linewidth}{!}{%
    \begin{tabular}{@{}lcccc@{}}
    \thickhline
    \multicolumn{1}{c}{\multirow{2}{*}{\textbf{Methods}}} & \multicolumn{4}{c}{\textbf{HardBLINK}} \\ \cmidrule(l){2-5} 
    \multicolumn{1}{c}{} & \(\mathbf{3}\) & \(\mathbf{4}\) & \(\mathbf{5}\) & \textbf{Avg.} \\ \midrule
    \fire~Aurora~\citep{bigverdi2025perception} & \(66.90\) & \(60.50\) & \(54.80\) & \(60.73\) \\
    \fire~ReVPT-\(7\)B~\citep{zhou2025reinforced} & \(68.55\) & \(55.65\) & \(60.48\) & \(61.56\) \\
    \snow~\begin{tabular}[c]{@{}l@{}}Visual Sketchpad~\citep{hu2024visual} \\ (GPT-5 Mini)\end{tabular} & \(71.77\) & \(62.90\) & \(56.45\) & \(63.71\) \\ \midrule
    \snow~GPT-5 Mini & \(62.10\) & \(53.23\) & \(41.49\) & \(52.42\) \\
    \snow~GPT-5 Mini + \textsc{P}\(^2\) & \cellcolor{rowcolor} \(\mathbf{82.26}\) & \cellcolor{rowcolor} \(\mathbf{87.90}\) & \cellcolor{rowcolor} \(\mathbf{74.19}\) & \cellcolor{rowcolor} \(\mathbf{81.45}\) \\ \midrule
    \snow~Gemini 2.5 Pro & \(66.13\) & \(55.65\) & \(45.16\) & \(55.65\) \\
    \snow~Gemini 2.5 Pro + \textsc{P}\(^2\) & \cellcolor{rowcolor} \(\mathbf{77.42}\) & \cellcolor{rowcolor} \(\mathbf{83.87}\) & \cellcolor{rowcolor} \(\mathbf{71.77}\) & \cellcolor{rowcolor} \(\mathbf{77.68}\) \\ \midrule
    \snow~InternVL3.5-\(2\)B & \(\mathbf{39.52}\) & \(28.23\) & \(20.97\) & \(29.57\) \\
    \snow~InternVL3.5-\(2\)B+\textsc{P}\(^2\) & \(35.48\) & \(\mathbf{29.03}\) & \(\mathbf{32.26}\) & \(\mathbf{32.25}\) \\ \midrule
    \snow~InternVL3.5-\(4\)B & \(55.65\) & \(36.29\) & \(29.84\) & \(40.59\) \\
    \snow~InternVL3.5-\(4\)B+\textsc{P}\(^2\) & \cellcolor{rowcolor} \(\mathbf{72.58}\) & \cellcolor{rowcolor} \(\mathbf{70.97}\) & \cellcolor{rowcolor} \(\mathbf{66.13}\) & \cellcolor{rowcolor} \(\mathbf{69.89}\) \\ \midrule
    \snow~Qwen3VL-\(4\)B & \(55.65\) & \(46.77\) & \(38.71\) & \(47.04\) \\
    \snow~Qwen3VL-\(4\)B+\textsc{P}\(^2\) & \cellcolor{rowcolor} \(\mathbf{73.39}\) & \cellcolor{rowcolor} \(\mathbf{61.29}\) & \cellcolor{rowcolor} \(\mathbf{48.39}\) & \cellcolor{rowcolor} \(\mathbf{61.02}\) \\ \thickhline
    \end{tabular}}
    \caption{\textbf{HardBLINK Breakdown.} We report accuracy (\%) on different sub-tasks in HardBLINK~\citep{bigverdi2025perception}: 3 point, 4 point and 5 point. Each setting is increasingly difficult from the prior, wherein more candidate points are presented to the MLLM.}
    \label{tab:hardblink_breakdown}
\end{table}

\begin{figure}
    \centering
    \includegraphics[width=0.9\linewidth]{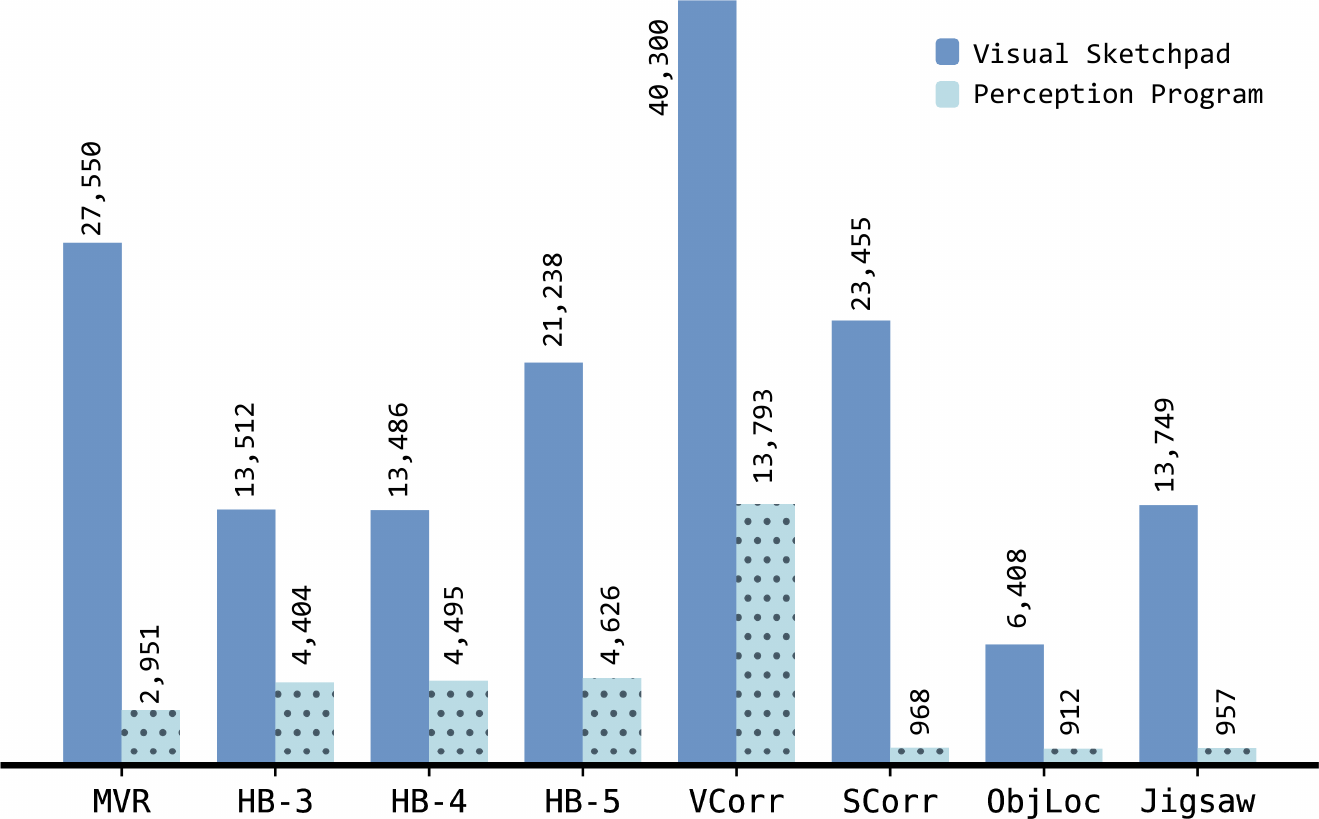}
    \caption{\textbf{Average Tokens/Sample.} Comparison of Visual Sketchpad (with GPT-5 Mini as LLM) and GPT-5 Mini with \textsc{P}\(^2\) on average token per sample across all six sub-tasks. \textsc{P}\(^2\) incurs significantly lower token cost.}
    \label{fig:tokusage}
\end{figure}

\section{Perception Program Details}
\label{appdx:pp}

\begin{figure}[!t]
    \centering
    \includegraphics[width=\linewidth]{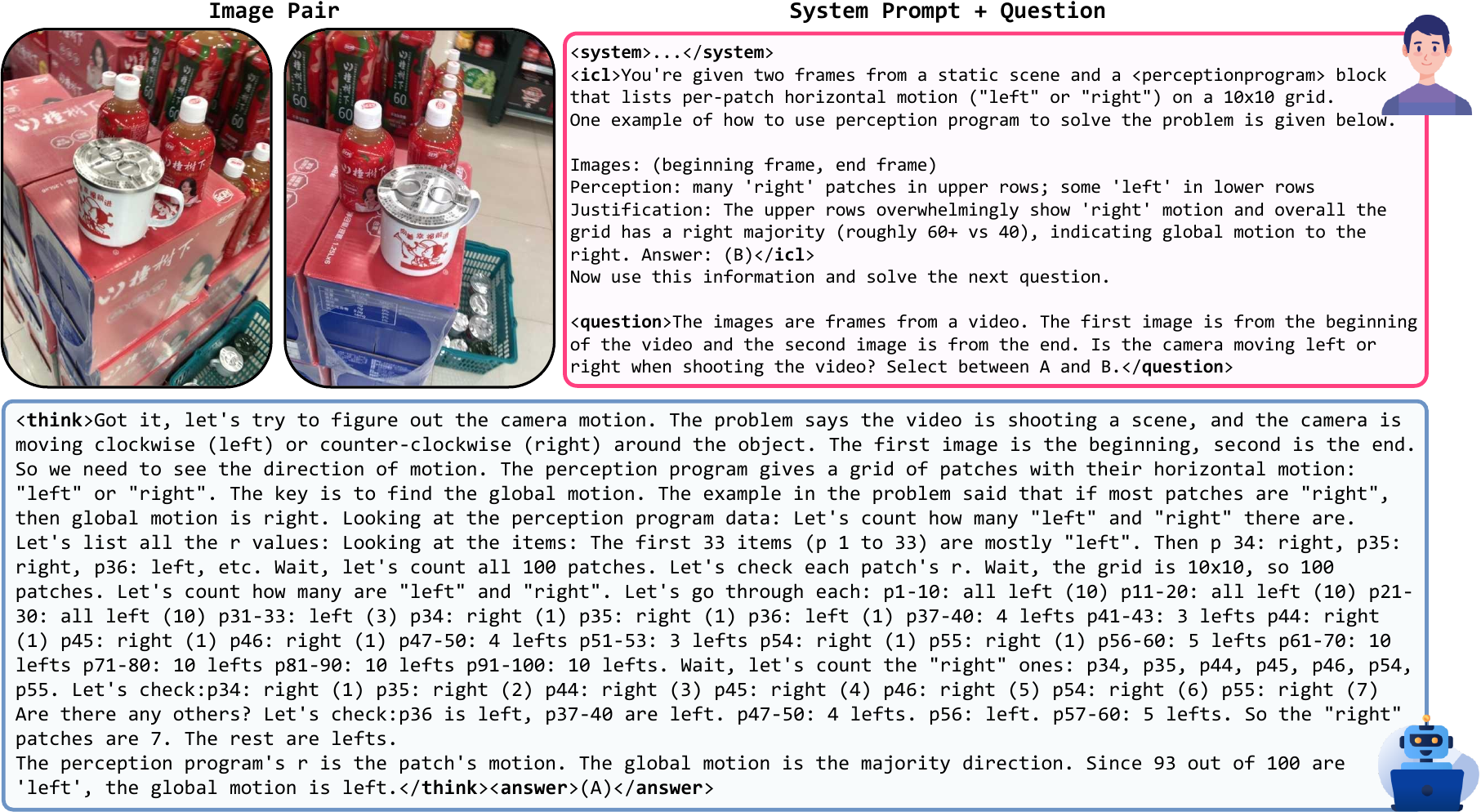}
    \caption{\textbf{Open-Source Prompt with \textsc{P}\(^2\) ICL.} We present a sample prompt for open-source MLLMs (e.g., Qwen3VL and InternVL3.5). We include a single in-context example describing the use of \textsc{P}\(^2\). Both Qwen3VL and InternVL3.5 reason with the given \textsc{P}\(^2\) to compute the \textcolor{correctgreen}{correct answer (A)} to the question.}
    \label{fig:opensource_prompt_pp}
\end{figure}

In this section, we discuss additional details about Perception Programs. Mainly, we provide samples of prompts for both frontier and open-source MLLMs. We also detail in-context (ICL) example that we use to query the open-source MLLMs. Recall that frontier models, GPT-5 Mini and Gemini 2.5 Pro, work as is and do not require any ICL examples. However, for both Qwen3VL and InternVL3.5, we provide a single in-context example as part of the system prompt, see~\cref{fig:opensource_prompt_pp} for an illustrative example for one question in multi-view reasoning task. 

\begin{figure*}[!t]
    \centering
    \includegraphics[width=\linewidth]{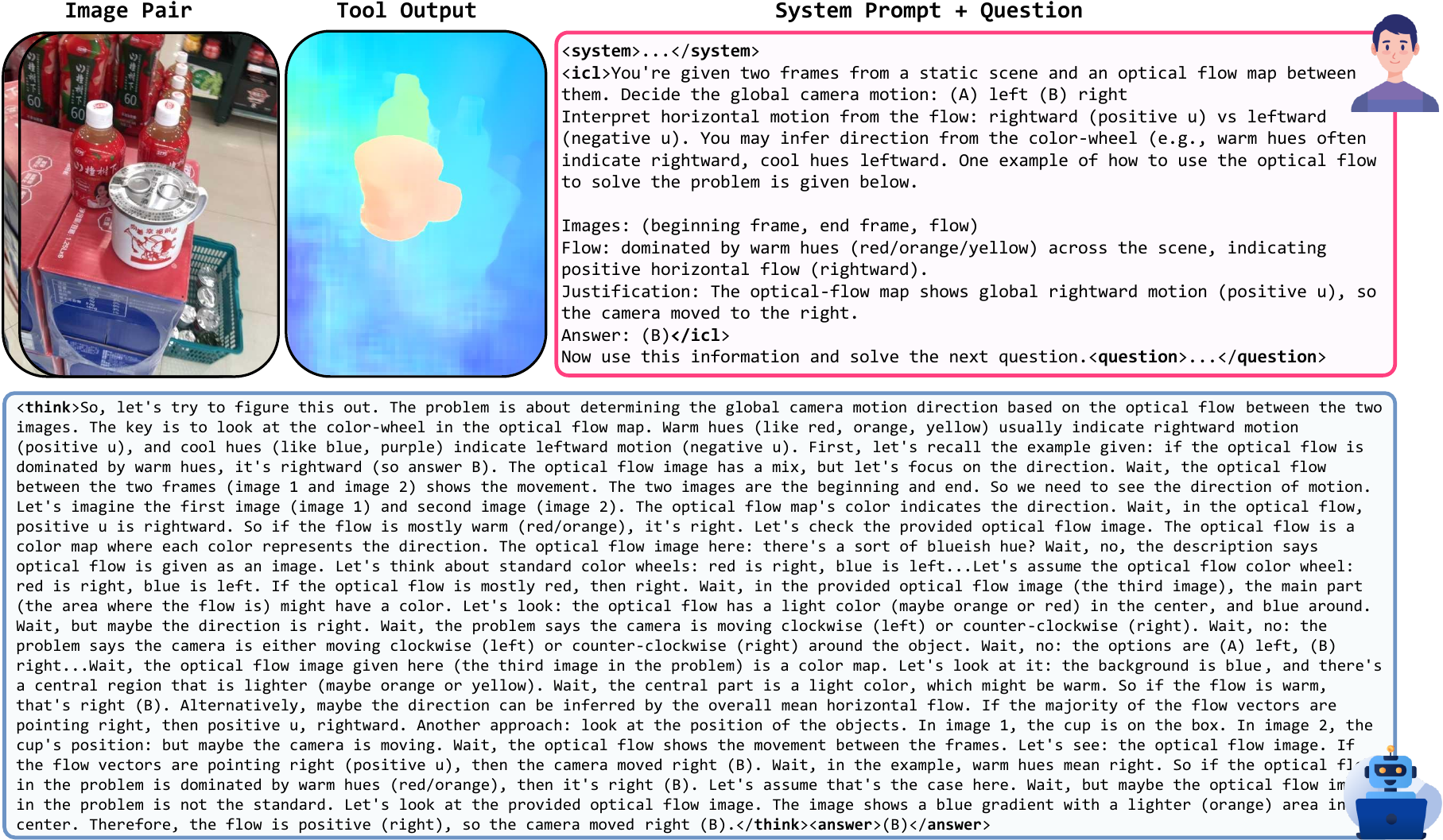}
    \caption{\textbf{Open-Source Prompt with Tool ICL.} We present a sample prompt for open-source MLLMs (e.g., Qwen3VL and InternVL3.5). We include a single in-context example describing the use of optical flow as tool output. Note how the example clearly illustrates that blue hues indicate left while warm hues indicate right motion, the MLLM (Qwen3VL in this example) concludes the same that flow is dominated by blue hues, yet gives the \textcolor{red}{wrong answer (B)}. Note that MLLM also uses a lot more tokens than its \textsc{P}\(^2\) counterpart (exhausting almost the entire \(8192\) token budget), we use \dots for brevity purposes in this illustrative figure.}
    \label{fig:opensource_prompt_tool}
\end{figure*}

Note that the \textsc{P}\(^2\) rationale as part of in-context sample is taken from GPT-5 Thinking~\citep{gpt5team}. We prompt GPT-5 Thinking with the question and its \textsc{P}\(^2\) and ask it to output a short rationale on how it uses \textsc{P}\(^2\) to compute the answer. We include this obtained rationale as an in-context example for open-source MLLMs. This procedure is similar for all the tasks from the BLINK benchmark we consider in this work, along with HardBLINK. Note that we do the same for raw tool setting and provide exhaustive descriptions of the tool in terms of how to use it to get to the answer, see~\cref{fig:opensource_prompt_tool}. 

An important distinction to note is for InternVL3.5 (both \(2\)B and \(4\)B variants), we also include additional instructions. In all problems, we direct it to to not copy the in-context example as is, along with some problem-specific orientations and clarification. For multi-view reasoning, we mention that clockwise and left are used interchangeably. For relative depth, we explain that the comparison of which point is closest is based on depth range and not coordinates. For semantic correspondence, we reiterate to use similarity scores and not the coordinates for comparison. In visual correspondence we emphasize not to directly use euclidean distance between coordinates in different images to conclude which point corresponds to REF. However, these clarifications are not necessary for Qwen3VL.

For closed-source frontier LLMs, such as GPT-5 Mini and Gemini 2.5 Pro, we just query with the question along with the tool or \textsc{P}\(^2\) and do not provide any example. The frontier LLMs are able to understand how to use \textsc{P}\(^2\) on their own.

\section{Additional Related Work}
\label{appdx:more_rl}

In this section, we give a non-comprehensive summary of methods from the related work, expanding on some that were briefly mentioned while also introducing additional ones. We additionally note that several prior state-of-the-art BLINK results were obtained by methods that do not rely on tools, which we also include here.

\subsection{Tool-Use through Program Synthesis}
Methods in this category leverage program generation, typically Python, to structure the model's reasoning process. Instead of reasoning purely in natural language, these approaches produce executable code that coordinates external vision modules to enable compositional and interpretable visual reasoning.

\begin{itemize}
    \item \textbf{VisProg} \citep{gupta2023visual}: A neuro-symbolic approach in which the model uses in-context learning to generate modular Python programs that call vision models and image-processing tools. Demonstrates strong performance in compositional VQA, reasoning over image pairs, object tagging, and language-guided image editing.

    \item \textbf{ViperGPT} \citep{suris2023vipergpt}: Reduces the burden on large MLLMs by equipping GPT-based models with an API of callable vision-related subroutines. The model generates Python programs executed on images or video, improving visual grounding and compositional question answering, including cases requiring external knowledge.

    \item \textbf{Thyme} \citep{zhang2025thyme}: Enhances logical reasoning by enabling models to perform image-level manipulations such as cropping, rotation, and contrast adjustment. Trained via a two-stage pipeline combining supervised fine-tuning and GRPO with adaptive temperature sampling.

    \item \textbf{MMFactory} \citep{fan2024mmfactory}: Addresses deployment challenges such as performance constraints and computational limits. Proposes a model that composes programmatic solutions from a tool repository while also suggesting metrics and benchmarks, taking user illiteracy into account to improve real-world usability.
\end{itemize}

\subsection{Tool-Use through Chain-of-Thought}
These methods integrate tool usage directly into the reasoning trajectory of the model, often allowing the system to call vision specialists or perform visual operations as part of its intermediate reasoning steps rather than through offline program synthesis.

\begin{itemize}
    \item \textbf{LATTE} \citep{ma2025latte}: Introduces \(8\)B vision-language models trained to incorporate outputs from multiple vision specialists as part of a think--act--observe reasoning loop. Supports tasks including object recognition, depth estimation, text extraction, and mathematical operations.

    \item \textbf{VisualSketchpad} \citep{hu2024visual}: A hybrid method that integrates code generation and tool usage into chain-of-thought reasoning. Inspired by sketch-based human problem solving and implemented on a GPT backbone.

    \item \textbf{VigoRL} \citep{sarch2025grounded}: Highlights the gap between the success of RL in math/coding and its limited impact on visually grounded tasks. Proposes grounding reasoning steps in image regions and enabling zoom operations to focus on visually relevant details.

    \item \textbf{ReVPT} \citep{zhou2025reinforced}: Uses a GRPO-based RL framework to train multimodal LLMs to reason with visual tools such as detection, zooming, edge analysis, and depth estimation. Shows notable improvements on perception-heavy benchmarks (e.g., KV-Bench, BLINK, MMVP, MMStar), with particularly strong gains for 2B-scale models.

    \item \textbf{Visual Perception Token} \citep{yu2025introducing}: Trains models to emit tool-calling tokens that enable selective invocation of external visual modules. Provides tools for region selection-then-zooming and for supplying enriched vision tokens from an auxiliary vision tower.
\end{itemize}

\subsection{Other Prominent Methods}
This group includes methods that achieve strong results in multimodal reasoning without relying primarily on explicit tool calls. Many provide alternative training paradigms or new datasets that improve perception or reasoning capabilities.

\begin{itemize}
    \item \textbf{TULIP} \citep{tang2025tulip}: Based on Llama-3.2-11B, this method addresses limitations of CLIP/SigLIP in detailed visual interpretation. Introduces contrastive and reconstruction objectives and provides a drop-in replacement vision tower, yielding improved BLINK performance.

    \item \textbf{PerceptionLM} \citep{cho2025perceptionlm}: Promotes transparency by avoiding reliance on closed-source vision model annotations. Constructs a fully open perception-language model (8B LLM + vision tower) that achieves strong BLINK performance without using tools.

    \item \textbf{Zebra-CoT} \citep{li2025zebra}: Tackles the scarcity of high-quality sketch/diagram reasoning data by releasing a new interleaved image--text dataset and training the Anole-7B model on it. Instead of using external tools, the model directly generates auxiliary images.

    \item \textbf{OVR} \citep{wei2025open}: Argues that prior RL approaches under-scale cognitive behavior training. Proposes a large two-stage RL paradigm achieving strong gains in mathematical reasoning and BLINK tasks.
\end{itemize}

All of the methods discussed above combine to form a wide-range of baselines we compare our proposed \textsc{P}\(^2\) to in~\Cref{tab:mainresults,tab:more_results_1}; also refer to~\cref{sec:exps}.

\section{Additional Experimental Details}
In~\cref{subsec:analysis} we discussed the quality of visual interpretation of current MLLMs. We expand the discussion on on visual correspondence task and describe the two baselines we included, namely Naive and Oracle, also see~\cref{fig:vcorr_recon_ablation}.

\subsection{Additional Details on Visual Correspondence}

When evaluating the reconstructed \textsc{P}\(^2\) results, we considered the naive Euclidean and oracle baselines. We describe their setup as follows.

\textbf{Naive.} It receives ground-truth coordinates corresponding to the BLINK alternatives: REF in the reference image and A, B, C, D, E in the target image. It simply gives the answer as the point whose coordinate is closest to REF in the normalized coordinate space. This method completely disregards the visual content of the image and is therefore unsuitable to solve the task of visual correspondence. Its performance of 85\% indicates that most pairs of images indeed have low camera movement, which we later confirm in \cref{sub:app_disp_distr}.

\textbf{Oracle.} An implementation of correct usage of the \textsc{P}\(^2\) to navigate through correspondences is what we term as an oracle. Concretely, we first store the reference (REF) point coordinates. We then scan the correspondence \textsc{P}\(^2\) to find all the candidate points and read their `\(c\)' coordinates. For each candidate, we compute its Euclidean distance to the reference point and select the neighbor with the smallest distance. We then take this neighbor's `\(r\)' coordinate as the mapped position in the second image. Finally, we compare this mapped location with the coordinates of alternatives A, B, C, D, E, and choose the point whose coordinates are closest in Euclidean distance as the correspondence of the original reference point.


\begin{figure}
    \centering
    \includegraphics[width=\linewidth]{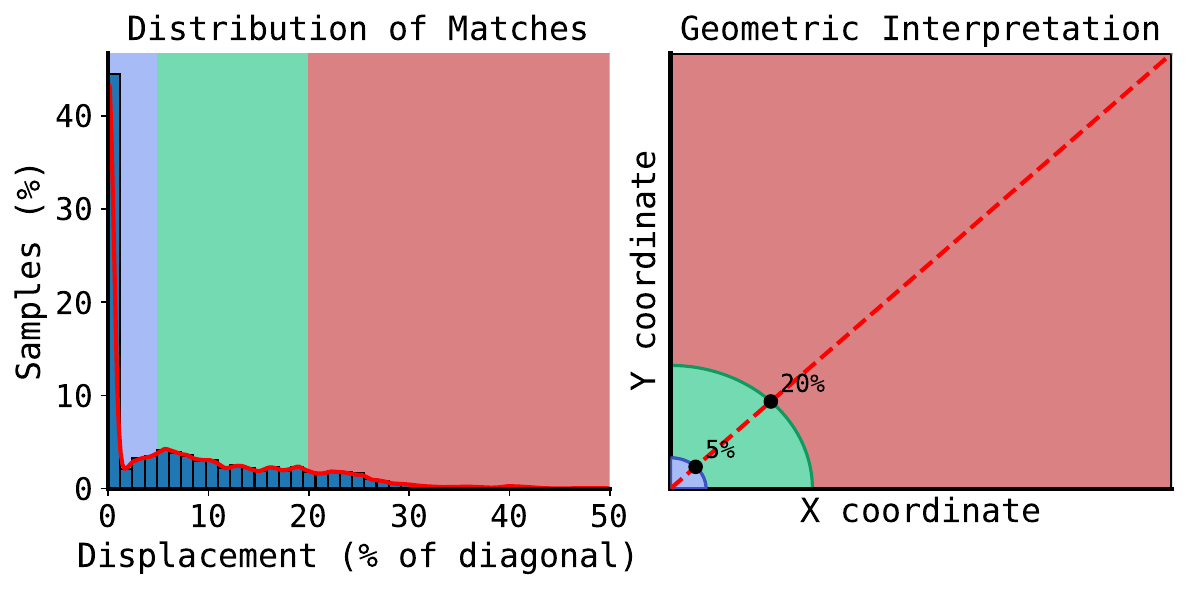}
    \caption{\textbf{Correspondence Distribution.} Illustration of distribution of correspondence markers in the visual correspondence task from BLINK validation set.}
    \label{fig:viscorr_distribution}
\end{figure}

\subsection{Distribution of Displacements}
\label{sub:app_disp_distr}
We here plot the distribution of visual correspondences in the BLINK validation set. The left pane illustrates the histogram and density of LoFTR~\citep{sun2021loftr} displacements across the whole dataset as a percentage of the diagonal. To give the reader a rough visual reference of the displacement ranges, we illustrate regions closer than 5\% (blue) and 20\% (green) of the diagonal, considering the normalized coordinate space. Colors between panes correspond to matching regions. We can see that the majority of displacements are closer than 5\% of image diagonal, further evidencing the fact that this dataset is biased towards low displacement between images.

\subsection{Breakdown on HardBLINK}
In~\cref{tab:hardblink_breakdown}, we present results on each sub-task, 3-, 4-, and 5-point, in HardBLINK benchmark introduced in~\citet{bigverdi2025perception} to complement the results presented in~\cref{tab:mainresults} (the HardBLINK performance reported there is the average of these sub-tasks).

Across all three HardBLINK settings, we observe a consistent trend: performance drops as the number of candidate points increases, but \textsc{P}\(^2\) substantially narrows this gap. Specialized baselines such as Aurora~\cite{bigverdi2025perception}, ReVPT~\citep{zhou2025reinforced}, and Visual Sketchpad (GPT-5 Mini)~\citep{hu2024visual} achieve average accuracies between \(60.73\%\) and \(63.71\%\), with modest degradation from \(3\)-point to \(5\)-point tasks. In contrast, raw MLLMs struggle more severely as difficulty increases, i.e., GPT-5 Mini falls from \(62.10\%\) on 3-point to \(41.49\%\) on 5-point sub-task. Both GPT-5 Mini and Gemini 2.5 Pro immensely benefit from \textsc{P}\(^2\), and even the smaller InternVL3.5-\(4\)B and Qwen3VL-\(4\)B observe \(+29.30\%\) and \(+13.98\%\) increase. 

Overall, \textsc{P}\(^2\) not only lifts all base models, but is particularly effective in the more challenging 4- and 5-point regimes, where raw MLLMs otherwise collapse.

\section{LLM Usage Statement}
\label{appdx:llm_usage}

In this manuscript, we used several MLLMs as part of our experimental setup and we have described the necessary details in~\cref{sec:exps,appdx:pp}. Other than that, we also used LLMs (ChatGPT) to help with refining the manuscript in terms of fixing grammatical errors in writing and with plotting codes for various figures. The authors did not use any LLM in any part of ideation, experimental design, analysis of results, and implementation of core methodology.

\end{document}